\documentclass[preprint,12pt]{elsarticle}

\usepackage{amssymb}
\usepackage{amsmath}

\usepackage{algorithm}
\usepackage{algorithmic}
\usepackage{soul}
\usepackage{ulem}
\usepackage{xcolor}
\usepackage{xspace}

\newcommand{\ours}{cAPM\xspace}

\journal{Computers in Biology and Medicine}

\begin{document}

\begin{frontmatter}



\title{\ours: Continual AI-Assisted Pace-Mapping \\with Active Learning}


\author[1]{Dylan O'Hara\fnref{cofirst}}
\ead{dbo8671@rit.edu}
\author[1]{Pradeep Bajracharya\fnref{cofirst}}
\author[1]{Casey Meisenzahl}
\author[2,3,4,5]{Karli Gillette}
\author[4,5]{Anton J. Prassl}
\author[4,5]{Gernot Plank}
\author[6]{Saman Nazarian}
\author[7]{Roderick Tung}
\author[8]{John L Sapp}
\author[1]{Linwei Wang}
\ead{linwei.wang@rit.edu}
\fntext[cofirst]{These authors contributed equally as co-first authors.}
\address[1]{Rochester Institute of Technology, Rochester, New York, USA}
\address[2]{Department of Biomedical Engineering, University of Utah, Salt Lake City, Utah, USA }
\address[3]{Scientific Computing and Imaging Institute, University of Utah, Salt Lake City, Utah, USA }
\address[4]{Gottfried Schatz Research Center, Division of Medical Physics and Biophysics, Medical University of Graz, Graz, Austria}
\address[5]{BioTechMed-Graz, Graz, Austria}
\address[6]{University of Pennsylvania Perelman School of Medicine, Philadelphia, Pennsylvania, USA}
\address[7]{The University of Arizona College of Medicine, Phoenix, Arizona, USA}
\address[8]{QEII Health Sciences Centre, Dalhousie University, Halifax, Nova Scotia, Canada}

\begin{abstract}
Ventricular tachycardia (VT) is a life-threatening rhythm disorder and a major cause of sudden cardiac death. Pace-mapping is a standard clinical procedure for identifying the intervention target during catheter ablation of VT. It requires the clinicians to pace different sites in the ventricles and rapidly interpret the resulting electrocardiograms (ECGs) to determine where to pace next or whether a target site has been identified. Active learning-based AI models have been proposed to guide clinicians to the next pacing site, showing promise in reducing the number of pacing sites and improving the efficiency of pace-mapping. Existing methods, however, require retraining from scratch for each target localization without the ability to transfer knowledge across multiple VTs within the same patient or across patients. We introduce \ours for continuous AI-assisted pace-mapping to continually capture and transfer knowledge accumulated from past pace-mapping data to reduce the number of pace-mapping data needed for future target VTs. This is made possible by a task-agnostic surrogate neural network that learns the mapping from pacing sites to 12-lead ECG morphology, an active-learning strategy that refines this surrogate model by progressively selecting the most informative pacing site for each target, and a continual learning strategy to do so sequentially while retaining knowledge from prior targets. Evaluated on an \textit{in-silico} testbed consisting of sequentially-presented localization tasks across different physiological conditions and ventricular geometries, \ours with and without replay of past data samples on average achieved an approximate $81\%$ probability of localizing within clinical tolerance ($5$ mm accuracy) using just $4.5$ pace-mapping sites, compared to the state-of-the-art active-lerning method achieving $38\%$ probability using $13.7$ pacing sites. These results provide a strong basis for preparing \ours towards prospective \textit{in-vivo} preclinical and clinical studies where it can be used to guide pace-mapping on the fly.
\end{abstract}

\begin{graphicalabstract}
\end{graphicalabstract}



\end{frontmatter}



\section{Introduction}
\label{sec_introduction} 
Ventricular tachycardia (VT) is a life-threatening arrhythmia caused by irregular, rapid electrical activation in the ventricles, the lower chamber of the heart \cite{samuel2022ventricular}. Such abnormal activity disrupts effective blood circulation to the rest of the body and can result in sudden cardiac death \cite{koplan2009ventricular}. To treat VT, clinicians must identify the source of this abnormal activation and use treatment methods such as catheter ablation to destroy the target \cite{meisenzahl2024boatmap}. A widely used clinical practice for localization, known as \textit{pace-mapping}, involves electrically stimulating different sites of the ventricles and comparing the induced electrocardiogram (ECG with that from the target VT \cite{josephson2005using}. When the QRS component of the induced ECG matches maximally with that of the target VT, the corresponding pacing site is considered as the target for ablation \cite{josephson2005using}. While effective, this procedure is invasive, labor-intensive, and reliant on expert cardiologists' rapid interpretation of the paced-ECG signals during the pace-mapping process.

To reduce the trial-and-error nature of pace-mapping, deep learning (DL) and machine learning (ML) approaches can be used to provide guidance about the location of the VT origin from 12-lead ECGs. Existing approaches can be broadly categorized into population-based and patient-specific methods. Population-based approaches \cite{yokokawa2012automated, gyawali2017automatic, pereira2019automated, chang2022high} train DL models on a large dataset of ECGs and pacing sites from a diverse cohort of patients \cite{meisenzahl2024boatmap}, and then apply the trained model to new patients. These methods assume that the new patients lie in the same distribution as the training cohort. However, even for ECGs originating in the same location within the ventricles, they can vary with many physiological and pathological conditions, such as heart and thorax geometries, patient-specific conduction properties, and disease-related remodeling of the ventricles \cite{gyawali2021learning}. Moreover, these methods typically require a large amount of 12-lead ECG data with their corresponding \textit{labels} of pacing locations, which are not trivial to acquire in clinical practice. 

In contrast, patient-specific approaches \cite{sapp2017real, zhou2019localization, missel2020hybrid} aim to train separate models for each individual patient \cite{meisenzahl2024boatmap}, thereby reducing the impact of inter-patient variability. While such methods can potentially improve personalization, a key challenge lies in the strategy for obtaining training data for individual patients, which requires pacing their ventricles from multiple cardiac sites and recording the corresponding ECGs. If the number of pace-mapping data needed for training is similar to \textit{un-guided} pace-mapping, it would defeat the model's intended purpose to improve the efficiency of pace-mapping.

To address these challenges, active learning (AL) has been proposed as a patient-specific solution for VT localization \cite{missel2020hybrid,meisenzahl2024boatmap}. AL is an ML paradigm that iteratively selects the most informative and uncertain data samples (in this case, pacing sites and their corresponding 12-lead ECGs) to train an ML model. In this context, a surrogate model, such as a Gaussian Process (GP), is used to capture the relationship between pacing sites and the target ECG using an error measure between the predicted and ground-truth ECG signal \cite{meisenzahl2024boatmap}. Minimizing the GP-approximated error measure when tending to its uncertainty estimates, through the optimization of an acquisition function (AF), balances the exploitation and exploration of the search space to identify the next pacing site. In doing so, AL minimizes the number of pace-mapping sites needed to achieve accurate localization for each target VT, improving the efficiency of the pace-mapping procedure \cite{missel2020hybrid,meisenzahl2024boatmap}.

A key limitation of the existing approach, however, is that the surrogate model is designed to estimate an error measure between a target ECG and any other paced ECG, and is therefore tied specifically to the ECG signal of a target VT. As a result, to localize each target VT (which we refer to as \textit{tasks} hereafter) requires running the AL procedure from scratch. This means collecting training data and developing the surrogates separately for each VT, even for multiple VTs on the same heart or across patients with similar conditions. This hinders the transfer of knowledge that is already gained from past pace-mapping data, limiting its applicability in clinical use.

In this work, we overcome the above limitations with a novel \textit{continual AL} framework for VT localization that can continually capture and transfer knowledge accumulated from past pacemapping data, reducing the number of pace-mapping data needed for future target VTs and thereby continually improving the efficiency of localizing VT origins. We refer to this framework as \textit{continual AI-assisted pace mapping} (\ours), which includes three key innovations to existing AL methods in VT localization:
\begin{itemize}
    \item \textbf{Task-agnostic surrogate models: }Instead of approximating a task-specific function tied to a single target ECG, we design reusable, general-purpose neural network surrogates that \textit{learn the mapping from pacing sites to ECG signals.} Contrasting prior approaches that model the error with respect to a target ECG, this general-purpose surrogate can be continuously refined across VT morphologies and subjects' physiological/pathological conditions, enabling continual accumulation and transferring of knowledge from past data to future pace-mapping tasks.

    \item \textbf{Active learning integration}: We embed this surrogate with an AL strategy that leverages the model's uncertainty estimates to guide sampling. By iteratively selecting the most informative pacing sites, the surrogate can be refined efficiently with only a small number of labeled examples per VT target per patient.

    \item \textbf{Continual learning across VTs:} 
    Finally, to handle the sequential nature of clinical practice, where multiple VT morphologies per patient and multiple patients are encountered over time, we use a continual learning strategy to allow the surrogate to adapt to new tasks while retaining knowledge from prior ones, thereby enabling efficient transfer of information both within and across patients. More specifically, we consider two alternative strategies to address catastrophic forgetting with the motivating difference being access to data from previous tasks. In one approach, we assume the use of a memory buffer of past data samples in order to learn how to quickly adapt a meta-neural surrogate to the specific task at hand (\ours-Meta). Alternatively, when learning across patients, it may be impractical to store and transfer past data samples directly: in this context, we consider an ensemble of past task-specific neural surrogates to generate aggregated predictions for the task at hand (\ours-Ensemble).  We evaluate the performance, memory footprint, and computation overhead of these two approaches.

\end{itemize}

\ours can be seen as bridging the gap between DL-based approaches and AL-based approaches. DL approaches are generalizable but reliant on large labeled dataset, while AL approaches are efficient, but patient-specific and non-transferrable. By combining continual learning with AL for a task-agnostic neural surrogate, the proposed framework leverages and accumulates past knowledge to reduce the number of pace-mapping sites needed for a new VT localization task, both within and across patients, ultimately improving the efficiency and clinical utility of AI-assisted pace-mapping.

Because of the AL nature of \ours, which requires data collection on-the-fly, it is difficult to evaluate with retrospective \textit{in-vivo} data since the latter tend to have limited coverage of the ventricles to support acquisition of data as needed \cite{meisenzahl2024boatmap}. We thus leverage high-fidelity computer simulations as our test bed to ensure that data are always available where they need to be acquired. To this end, we evaluated \ours on two distinct heart geometries, each considering five different tissue conditions, \textit{i.e.}, healthy myocardium and four settings with myocardial scars with varying sizes and locations. To assess the robustness and adaptability of our approach, we designed three setting configurations with progressively increasing difficulty: (i) localization of different pacing sites within the same ventricular geometry and physiological condition, (ii) localization within the same geometry but under different physiological conditions, and (iii) localization across different geometries and tissue conditions. Each setting was organized as a continuous sequence of localization tasks, in which the model incrementally adapts its knowledge from previous tasks to subsequent ones. For each configuration, we constructed 5 randomized 12-task sequences and repeated the evaluation over 10 random seeds. We compared both of our presented methods against our previously-published GP-based AL method for VT localization, known as the \textit{BOATMAP} \cite{meisenzahl2024boatmap}, across all settings. Evaluation metrics considered the number of pacing sites needed to reach the target site, and the localization accuracy of the target site measured in terms of error between the target and the final predicted pacing site.

Our experimental results showed that across all tasks and settings both \ours-Meta and \ours-Ensemble reached pacing site accuracy in significantly fewer steps ($4.1 \pm 2.0$ sites and $4.8 \pm 1.7$ sites respectively) compared to our previous technique \textit{BOATMAP} ($13.7 \pm 3.5$ sites), and as well showed improved reliability in reaching the clinical accuracy threshold, $92\%$ and $65\%$ respectively, compared to $38\%$. These empirical results, while \textit{in-silico}, provided a strong basis to prepare \ours towards prospective pre-clinical and clinical \textit{in-vivo} validations of \ours as a next step of its clinical translation.

\section{Problem Formulation} \label{problem_formulation}
The task of VT localization can be framed as follows: given the QRS complex of a 12-lead ECG recorded during VT, denoted as $y_{target}$, the goal is to identify the pacing site in the three-dimensional ventricular space $x = (u,v,w)$ that best explains the observed ECG. This is typically achieved by comparing the QRS complexes of ECGs generated from stimulated pacing at candidate sites \(y_x\) with the target VT ECG, and selecting the site that maximizes a similarity metric such as the correlation coefficient (CC).  

\begin{eqnarray}
    x^{\ast} = argmax_{x \in X} \{CC(y_{target}, y_{x})\}
\end{eqnarray}

\noindent Here, \(X\) denotes the pacing coordinate space in the ventricles over which the search for the best candidate pacing site is carried out.  

In naive formulations,
a surrogate function can be fit to $\{CC(y_{target}, y_{x})\}$ via an AL approach \cite{meisenzahl2024boatmap}. Because $\{CC(y_{target}, y_{x})\}$ changes with $y_{target}$, the model must relearn the mapping between $x$ and the CC values from scratch for each new target VT. This ignores the inherent structure shared across tasks and discards previously acquired knowledge. In real-world scenarios, multiple subjects and multiple VT morphologies per subject are encountered in sequence. Without mechanisms for knowledge transfer, naive AL leads to redundant data collection, increased pacing requirements, and inefficiency in clinical workflows. Therefore, the central problem is to design an AL framework that not only approximates the pacing--ECG relationship efficiently for a given VT target, but also leverages knowledge across sequential tasks. In \ours, we formalize this by continually and actively learning a surrogate function that maps pacing locations to QRS complexes of ECG signals. 

\section{Method}
\begin{figure}[t]
  \centering
  \includegraphics[width=\linewidth]{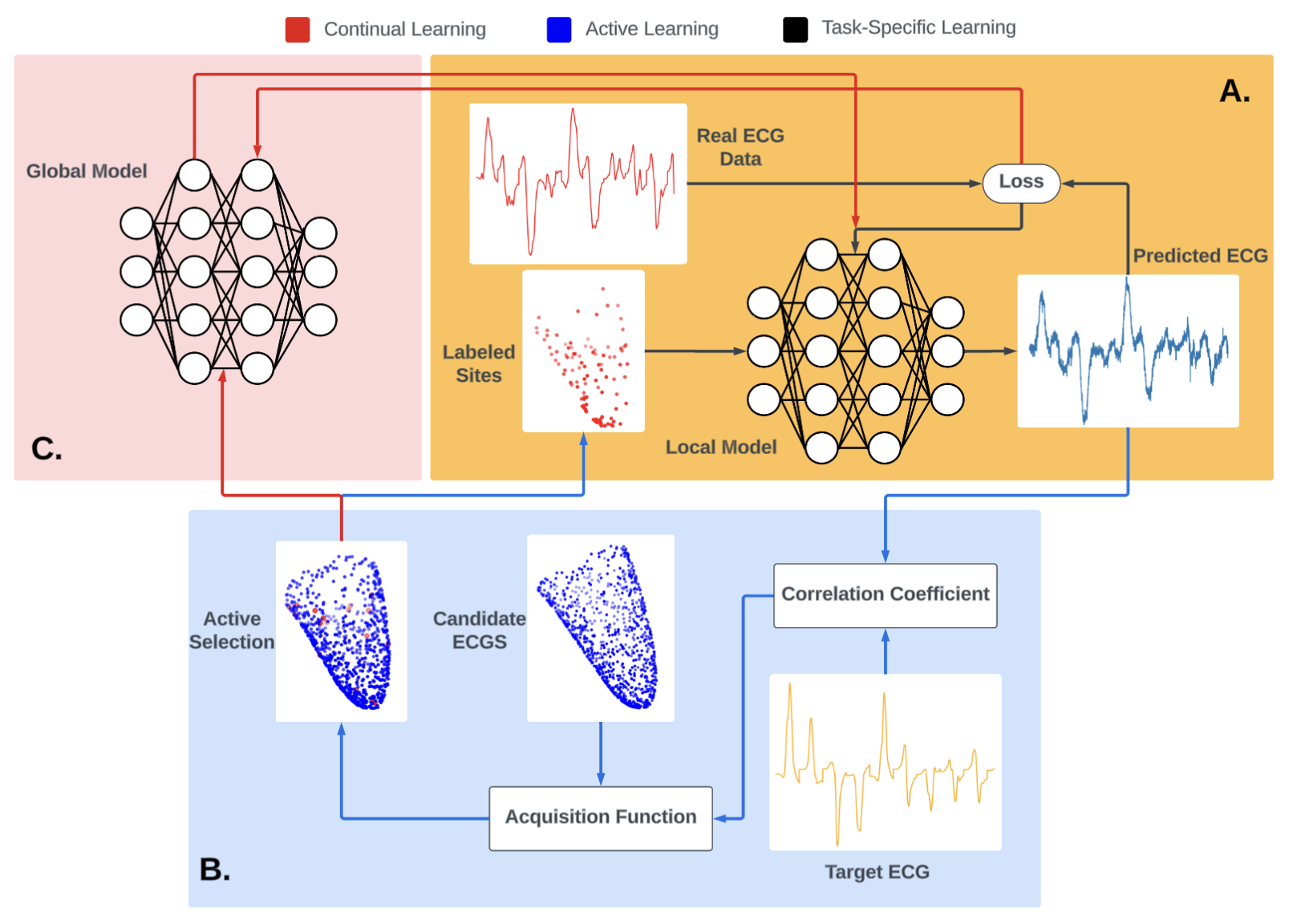}
  \caption{Overview of the \ours, which includes: A) a task-agnostic surrogate model that generates the QRS morphology of 12-lead ECG from a input pacing-site coordinate, B) an active learning approach to train this surrogate by intelligently guiding selection for the next pace-mapping site to query for 12-ECG data, 
  and C) a continual learning strategy to maintain the relation between a surrogate specific to the VT target and a "global" surrogate that can transfer knowledge across targets.}
  \label{fig:method_overview}
\end{figure}

As outlined in Fig.~\ref{fig:method_overview}, \ours includes three key components that together enable efficient and adaptive pace-mapping. 
First, we employ a neural network surrogate model $f_\theta: x \mapsto y$ that learns the mapping from pacing locations to ECG signals, providing both predictions and uncertainty estimates of correlation with the target VT morphology. Second, when localizing the origin of each target ECG, we introduce an uncertainty-driven AL strategy that improves the neural surrogate with active selection of patient-specific pace-mapping data, balancing exploration of unexplored regions with exploitation of promising sites to accelerate convergence toward the true origin. Finally, when localizing across target ECGs and even across hearts, we consider two different continual learning strategies.The first strategy (\ours-Meta) uses a continual meta-learning structure with an inner/outer optimization structure: the inner optimization fits the meta-model to pace-mapping data locally on each task, and the outer optimization accumulates the loss of these task-specific models to optimize the meta-model and thereby accumulates knowledge over tasks; to combat catastrophic forgetting across sequential tasks, a memory buffer is employed to maintain limited access to data from previous tasks. 
Alternatively, the second strategy (\ours-Ensemble) uses an ensemble of models such that only the newest model in the ensemble is trained on the current task while other models are frozen from their completed tasks, conserving information over time without the need for retaining data in memory buffers. Both work by transferring knowledge across sequential tasks, but do so using significantly different methods that can work with different practial constraints, in particular the possibility to retain past data samples for both memory and ethical concerns. Together, these three components form a unified framework that achieves continually accurate and data-efficient pacing site localization for target ECGs.

\subsection{Deep Neural Network Surrogates of Paced ECG Generations} \label{section_DNN}
To map the relation between the pacing sites \(x = (u, v, w)\) and the corresponding ECG signal \(y\), we choose a deep neural network (DNN) surrogate to model:
\begin{eqnarray}
    y = f_{\theta}(x) = f_{\theta}(u, v, w)
\end{eqnarray}
where $f_\theta$ is 
a multilayer perceptron (MLP) architecture with 
parameter $\theta$, optimized by the mean squared error loss between the predicted and true $y$ at each paced location.

To extract uncertainty about the 
predicted signal \(y\), 
we extract its high-frequency noise component $s_n$ using a band-pass filter (HPF) designed for the noise frequency range of \(40Hz\) to \(150Hz\) and calculate its standard deviation \(\sigma_n\) as:
\begin{eqnarray}
s_n = \text{HPF}(y, 40-150 \textit{Hz}), \quad 
    \sigma_n = \textrm{standard\_deviation}(s_n)
\end{eqnarray}

To evaluate the CC of this predicted $y$ with the target $y_\text{target}$ when including uncertainty measures, 
we create a \textit{clean} prediction $y_d$
by subtracting the noise $s_n$ from $y$ followed by smoothing with a Savitzky–Golay filter as:
\begin{eqnarray}
     y_d = \textrm{SGFilter}(y - s_n)
\end{eqnarray}
along with two perturbed versions of $y$ as 
$y \pm \sigma_n$. 
We use these to define the mean and standard deviation of the CC measures between predicted $y$ and the target $y_\textit{target}$ as:
\begin{eqnarray} \label{cc_eqn}
    \mu(x) &= CC(y_d, y_{target}) \nonumber \\
    CC_{lower} &= CC(y_d +  \sigma_n, y_{target}) \nonumber \\
    \sigma(x) &= CC - 
    CC_{lower} \nonumber 
\end{eqnarray}

This formulation provides both an estimate of the expected similarity between the surrogate-predicted ECG and the target and an uncertainty measure reflecting the impact of residual noise in the prediction.

\subsection{Active Learning of DNN Surrogates} \label{section_al}

We actively train $f_\theta$ with pace-mapping sites selected by the expected improvement (EI) criterion to balance exploration of uncertain regions with exploitation of regions likely to yield high similarity:

\begin{eqnarray}
    \mathrm{EI(}x) & = (\mu(x)-\mu^+)\mathrm{\Phi}\bigg(\frac{\mu(x)-\mu^\textrm{+}}{\sigma(x)}\bigg) + \sigma(x){\phi}\bigg(\frac{\mu(x)-\mu^\textrm{+}}{\sigma(x)}\bigg),
\label{eq:ei}
\end{eqnarray}

where \(\mu(x)\) and \(\sigma(x)\) are described in Eqn~\ref{cc_eqn}. \(\mu^+\) is the best observed CC so far, and \(\Phi\) and \(\phi\) denote the cumulative distribution function (CDF) and probability density function (PDF) of the standard normal distribution, respectively. By iteratively maximizing EI, the algorithm selects pacing sites that are either expected to improve the CC with the target ECG, or reduce uncertainty in regions where $f_\theta$ was not well approximated, ultimately enhancing the efficiency and accuracy of the pace-mapping process.

On each task (target VT), the active learning occurs in an interacitve process. In each iteration, using the current surrogate $f_\theta$, new ECG data are acquired on the suggested pace-mapping site $x$. The newly acquired data are then used to update the neural surrogate model. The process is repeated until the correlation coefficient between the predicted ECG and the target ECG is greater than or equal to a predefined threshold $\mathbf{\epsilon}$ determined by the accuracy requirement of the localization procedure. 

\subsection{\ours-Meta: Continual Active Learning of a Meta Surrogate with Exempler Replay} \label{section_metalearning}
To enable $f_\theta$ to be learned continually across multiple target VTs and patients, we introduce \ours-Meta that leverages continual meta-learning via \textit{meta-experience replay} (MER) \cite{riemer2018learning} to maintain a meta-model of $f_\theta$ that can be quickly adapted to a specific target VT task with its actively acquired data samples. This meta-model is optimized by \textit{gradient aligment} from all current and past tasks, utilizing a memory buffer of representative samples from past tasks.

\subsubsection{Memory Initialization} 
We maintain a compact memory buffer $M$ of fixed size $m$, initially empty. After each active learning iteration in each task, memory $M$ is updated to incorporate newly acquired labeled data using reservoir sampling: considering a \textit{age} variable $g$ that records the total number of memory updates performed, a new sample will replace a reservoir sample with a probability $\frac{n}{g}$ with each reservoir sample has an equal $\frac{1}{g}$ probability to be replaced. This mechanism allows us to update the memory with new data sampling, ensuring that the surrogate remains effective across sequentially presented VT targets and patients. 

\begin{algorithm}[t]
\caption{\ours-Meta for task $h \in H$}
\label{algorithm_meta_learn}
\begin{algorithmic}[1]
    \REQUIRE Memory buffer $M$, meta model $f_\theta$, age $g$, target ECG $y_{\text{target}}$, candidate data $z_h$, learning rate $\alpha$, meta learning rate $\beta$
    
    \REPEAT
        \STATE $(x_{\text{new}}, y_{\text{new}}) \leftarrow \text{ActiveLearning}(f_\theta, z_h)$
        \STATE $\mathcal{\theta}_{\text{prior}} \gets \text{copy}(f_\theta)$
        \FOR{s in steps} 
        \STATE $\theta_{\text{post}} \gets \text{copy}(\theta_{\text{prior}})$
        \STATE $L_h = (x_{\text{new}}, y_{\text{new}}) \cup \text{shuffle}(M)$
        \FOR{e in epochs}
            \STATE $\theta_{{\text{post}}} \gets \theta_{\text{post}} - \alpha \cdot \nabla_{\theta_{\text{post}}} \text{Loss}(L_h)$
        \ENDFOR
        \STATE $\theta_{\text{prior}} \gets \theta_{\text{prior}} + \beta \cdot (\theta_{\text{post}} - \theta_{\text{prior}})$
        \ENDFOR
        \STATE $f_{\theta} \gets \theta_{\text{prior}}$
        \STATE \text{MemoryUpdate}($M, (x_{\text{new}}, y_{\text{new}}), g$)
    \UNTIL{$\text{CC}(y_{\text{target}}, y_{\text{new}}) > 0.97$}
\end{algorithmic}
\end{algorithm}

\subsubsection{Model Updates}
Algorithm~\ref{algorithm_meta_learn} outlines the integration of meta-learning with the active learning framework described  earlier. For each task \(h \in H\), the memory buffer \(M\) and the meta-model \(f_\theta\) are carried forward to enable continual learning across iterations. At the beginning of each active learning cycle, the meta-model  \(\theta_{\text{prior}}\) is recorded. The active learning method as described in Section \ref{section_al} is used to identify the next pace-mapping site to collect ECG. Once the new data point \((x_{\text{new}}, y_{\text{new}})\) is acquired, it is combined with samples from the memory buffer to train \(\theta_{\text{post}}\) from a copy of \(\theta_{\text{prior}}\). The meta-model parameter \(\theta_{\text{prior}}\) is then adapted using the difference between \(\theta_{\text{post}}\) and \(\theta_{\text{prior}}\), and given to \(f_{\theta}\) thus consolidating knowledge from both the newly acquired and previously seen tasks. The memory buffer is subsequently refreshed as described above. This continual active learning method  enables progressive adaptation and continual refinement of the DNN surrogate $f_\theta$ to be quickly adaptive to new tasks.

\subsection{\ours-Ensemble: Continual Active Learning with Grown Ensembles}
Because of the need to maintain a buffer of past samples, \ours-Meta may not apply in scenarios where access to past data are not possible due to logistical, privacy, or hardware constraints. To mitigate this issue, we introduce the alternative of \ours-Ensemble that dynamically assembles an ensemble of base learners $f_\theta$'s for each task. The method balances plasticity and stability using a learned weighting module, or "weighter", that selects and combines prior learners based on task performance and historical utility.

\subsubsection{Ensemble Weighting}
A separate neural network with parameter $\phi$, $w_\phi(x)$, is used to weight both the current and prior learners based on their performance in the current task. The weighter is implemented following a traditional probabilistic modeling structure, where the input is transformed by several learning layers followed by a final layer to turn logits into probabilities for each prior learner. 

The weighter starts with a significantly disruptive dropout rate of $p=0.5$ at the beginning of each task, forcing it to consider many models as possible sources of prior information. As the task progresses the dropout rate decreases by $0.05$ each round, allowing the model to give more weight to the best performing model. The final prediction is given as the weighted ($w_{\phi}$) average of each base learner ($f_{\theta_j}$) in the ensemble ($E$):

\begin{equation}
G(x) = \sum_{j=1}^{|E|} w_\phi(x)_j \cdot f_{\theta_j}(x)
\end{equation}

Using the dataset of the current task only, backpropagation is done to update the parameters of the weighter $w_\phi$ and the newest base learner $f_{\theta_{|E|}}$, while all other $f_{\theta_j} \in E$ remain unchanged.

\subsubsection{Growing and Pruning the Ensemble}
When a new task is started, an additional learner is appended to $E$ with its parameters randomly initialized, and its weight $w_{|E|}$ is $0$. When the size of $E$ exceeds a predefined value $K$, there will be a pruning stage to remove one of $f_{\theta_j}$'s in favor of a newly initialized $f_{\theta_{|E|}}$. To do this, for the $j$th prior learner, we keep a record of the ensemble weights at the convergent round for each task $w^h_j$ and compute its historical weight $\bar{w}_j^h$ as:
\begin{eqnarray}
    \bar{w}_j^h = \frac{1}{|w_j^h|} \sum_{i = 1}^{|E|} (w_j^h)_i, \quad
    w_j^h \text{ is the weight history of $f_{\theta_j}$} \nonumber
\end{eqnarray}
$f_{\theta_j}$ with the lowest historical weight $\bar{w}_j^h$ (where $j \leq K - k$ and we denote $k$ as the lenience) is removed from $E$. We put this restriction on $j$ to allow newer models to proliferate in a number of tasks before they can be decided to be of low weight.

\subsubsection{Integration within Continual Active Learning}
For the first task, we initialize the single $f_{\theta_1}$ using $l$ randomly selected pacing-sites. For each subsequent task, the first round of acquisition uses $w_\phi$ of the ultimate round in the previous task (after the ensemble has been pruned, if applicable). The full algorithm is shown in Algorithm \ref{algorithm_ensemble}.

\begin{algorithm}
\caption{\ours-Ensemble for task $h \in H$}
\label{algorithm_ensemble}
\begin{algorithmic}
    \REQUIRE ensemble $E$, weighter $w_{\phi}$, weight history $w^h$, base learner $f_{\theta}$, ensemble limit $K$, lenience $k$, target ECG $y_{\text{target}}$, candidate data $z_h$

    \REPEAT
        \STATE $(x_{\text{new}}, y_{\text{new}}) \leftarrow \text{Active Learning}(E, z_h)$
        \FOR{e in epochs}
            \STATE $w_{\phi} \gets w_{\phi} - \nabla_{w_{\phi}} \text{Loss}(X, Y)$
            \STATE $f_{\theta_{|E|}} \gets f_{\theta_{|E|}} - \nabla_{f_{\theta_{|E|}}} \text{Loss}(X, Y)$
        \ENDFOR
    \UNTIL{$\text{CC}(y_{\text{target}}, y_{\text{new}}) > 0.97$}
    \STATE $E \gets \text{Grow}(E, f_{\theta})$
    \IF{$|E| > K$}
    \STATE $E \gets \text{Prune}(E, w^h, k)$
    \ENDIF
    
\end{algorithmic}
\end{algorithm}

\section{Experiment and Results}
Unlike testing models trained \textit{offline}, the testing of \ours requires an ability to acquire paced ECG signals at locations where the AL component suggests. This raises significant challenges with \textit{in-vivo} testing, as retrospective \textit{in-vivo} pace-mapping data often are too sparse in coverage to provid data as requested by the AL method \cite{missel2020hybrid}. To build sufficient validation evidence towards prospective \textit{in-vo} evaluations,  we adopted the virtual testbed as described in \cite{meisenzahl2024boatmap}: leveraging high-fidelity ECG simulation that can be generated at a high spatial density, this testbed can support \ours's \textit{acquisition} of ECG data virtually anywhere on the ventricles. 

\subsection{Data}
\begin{figure}[t]
  \centering
  \includegraphics[width=\linewidth]{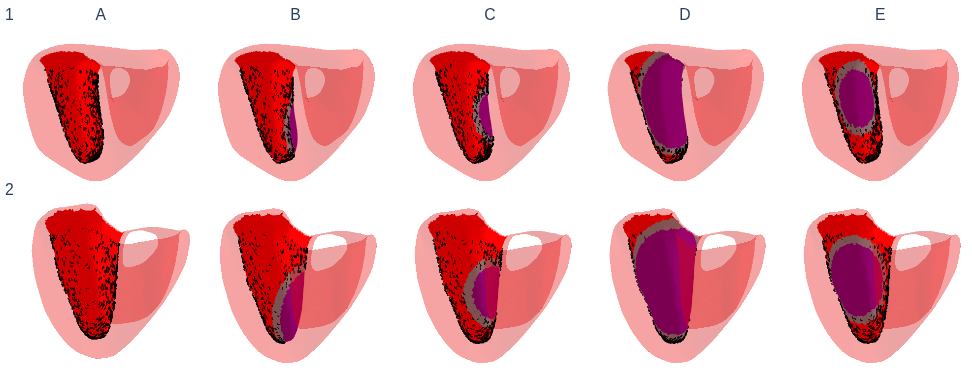}
  \caption{Ventricular geometry for two hears (one per row) and their respective different conditions (A: Healthy, B-E: Four different myocardial infarctions) used in the experiments. Purple represents scar, gray represents scar border, and black dots represents origin of activation at which 12 lead ECGs are simulated.}
  \label{fig_hearts}
\end{figure}

As detailed in \cite{meisenzahl2024boatmap}, this testbed included two human biventricular models obtained from the Experimental Data and Geometric Analysis Repository (EDGAR) \cite{aras2015experimental}. Each bi-ventricular model included five distinct settings of tissue properties, including a setting of healthy myocardium, and four settings of myocardial scars of various locations and sizes ranging between $1cm^2$ to $29 cm^2$ as illustrated in Figure~\ref{fig_hearts}. On each setting, 12-lead ECGs were simulated on densely-distributed pacing sites at an average resolution of 14 pacing sites per $cm^2$, where the average correlation coefficient (CC) between ECGs generated from pacing sites withing \(5mm\) is greater than 0.97: in other words, although the ECGs are precomputed offline, the pacing site selected for training will be on average only $1.5 mm$ away from what is suggested by \ours, ensuring the evaluation of \ours to not be confounded by the inability to acquire data where necessary.

\subsection{Baselines}
We evaluated our methods against the state-of-the-art AL-assisted pace-mapping approach, BOATMAP \cite{meisenzahl2024boatmap}, which is based on a GP-based surrogate model. This baseline showed considerable improvement over passive learning and random sampling methods, but required retraining to locate each target VT. In the worst case, we expect \ours to perform at least as well as an active GP-based strategy without continual learning.

For both cAPM methods, we established reasonable hyperparameters for our main experiments, with their effect ablated in Section \ref{subsec:ablation}. More specifically, for cAPM-Meta, we used a memory buffer size $|M| = 10$, inner model learning rate $\alpha = 0.001$, and outer model learning rate $\beta = 0.003$. Both the inner and outer models were implemented using the same base model $f_{\theta}$ which has $3$ layers of size $128$ each followed by ReLU activation functions, except for the last which outputs a prediction vector of size $12 \cdot 170 = 2040$. For cAPM-Ensemble, we chose the ensemble size limit $K = 10$, the lenience for prior models to remain in the ensemble is $k = 3$, and each prior model was implemented using the same $f_{\theta}$ as in the cAPM-Meta method. The weighter $w_{\phi}$ had a different MLP architecture: we used 5 layers, sizes $64, 64, 64, 32, 16$, with ReLU activation functions, except for the last which uses SoftMax to transform logits into a vector of probabilities (weights) for each prior model. The BOATMAP implementation was identical to the specification in \cite{meisenzahl2024boatmap}.

\subsection{Experiment Setup}
We designed three types of continual learning settings with increasing levels of difficulty. At the first level, target ECGs were drawn from the same heart under the same physiological condition, where only the pacing target varies. At the second level, target ECGs were drawn from the same ventricular geometry but across different underlying tissue properties. Finally, at the third level, both the geometry of the heart and the tissue properties varied simultaneously across tasks, testing the ability of \ours to adapt across heterogeneous patients and conditions.

For each of the three experimental settings, we tested on five different task sequences consisting of 12 sequential localization tasks each, where results on each task sequence were obtained from 10 different initialization/model seeds. For each localization task, a criterion of convergence success was set at when the ECG acquired at a suggested pacing site had a CC$\ge0.97$ with the target ECG, at which the pace-mapping was terminated. A pace-mapping was considered to be failed if it did not reach the convergence criteria at a maximum number of 25 iterations of acquisitions. 

\subsection{Evaluation Metrics}
We considered the following evaluation metrics:
\begin{itemize}
    \item Number of pace mapping sites: The number of pace mapping sites needed to reach a CC of \(97\%\) between the predicted ECG an the target ECG. This is calculated from cases where the target CC is reached within the maximum 25 number of pace-mapping sites. A lower number of pacing sites indicates a more efficient search. 
    \item Localization error: The Euclidean distance (in $mm$) between the actual target and the predicted pacing site. This is also calculated from cases where the target CC is reached within the maximum 25 number of pace-mapping sites. A smaller distance error indicates a more accurate localization of the origin of activation.
    \item Percentage of failed localizations at the maximum number of pacing sites allowed: Because the above two metrics only reflected the cases where the target CC of 97\% can be achieved, we also recorded the percentage of cases, for each 12-task sequence, when such termination criterion cannot be achieved within 25 pace-mapping sites. A higher percentage of failure indicates a less efficient search.
    
    \item Resource consumption of each method, including the time to finish a single task, the number of sampled data that is transferred across tasks, and the size of the saved model that is transferred across tasks.
\end{itemize}

\begin{figure}[t!]
  \centering
  \includegraphics[width=\linewidth]{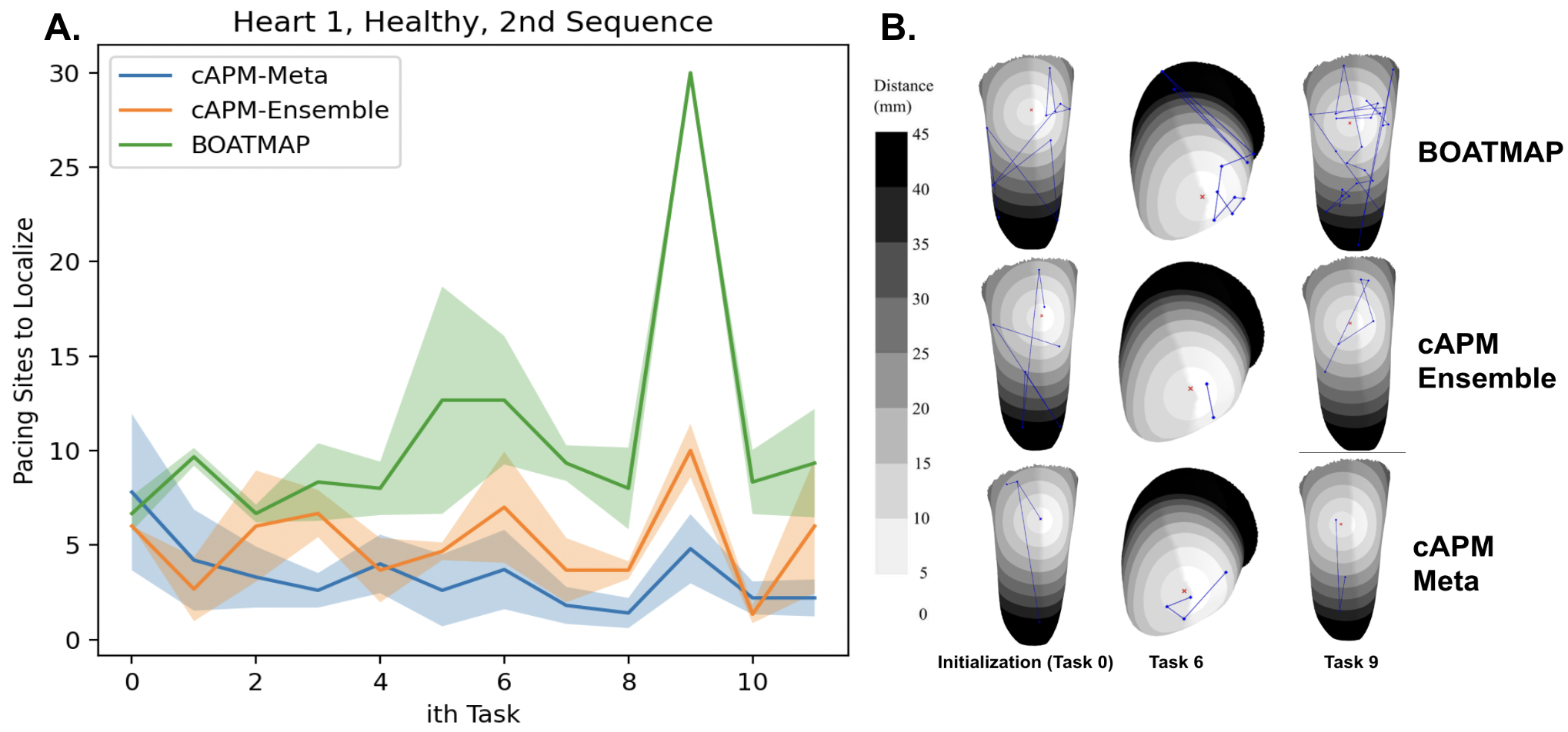}
  \caption{A: An example visualizing the number of pacing sites required by \ours-Meta, \ours-Ensemble, and BOATMAP to reach the convergence criteria on a sequence of 12 localization tasks on the same heart with healthy myocardium. Note that, after the first task, both \ours methods required only $3-6$ sites to reach the target in comparison to BOATMAP that required $10-15$ sites. Note also note \ours methods were able to do so in the task (task $\# 9$) that BOATMAP failed. B: Selected examples of localization tasks showing the number of pacing sites queried by BOATMAP \textit{versus} the two \ours methods.}
  \label{ssovertime}
\end{figure}

\subsection{Task Sequences across the Same Geometry and Physiological Conditions} \label{same_heart_same_condition}
In this set of experiments, we localized the pacing sites for different target ECGs on the same heart under identical physiological conditions, varying only the pacing target location. Clinically, this can be considered as attempting to localize multiple VTs on the same patient during the same procedure. 

\subsubsection{Localization Efficiency}
Figure~\ref{ssovertime} presents an example of the progression of all three models on a specific task sequence, showing the average number of pacing sites required for each of the 12 tasks where the shaded region represents the variability across initialization/model seeds. At the first task where all models started from scratch, all three method as expected performed similarly. As the sequence progressed, particuarly around the 4th task, BOATMAP continued to require 10-15 pacing sites to localize a target, whereas \ours required only 3-6 sites. Note that, at the 9th task where BOATMAP failed to converge to a solution, both \ours methods were able to leverage their acquired knowledge to converge using only a few pacing sites.

\begin{figure}[t!]
  \centering
  \includegraphics[width=\linewidth]{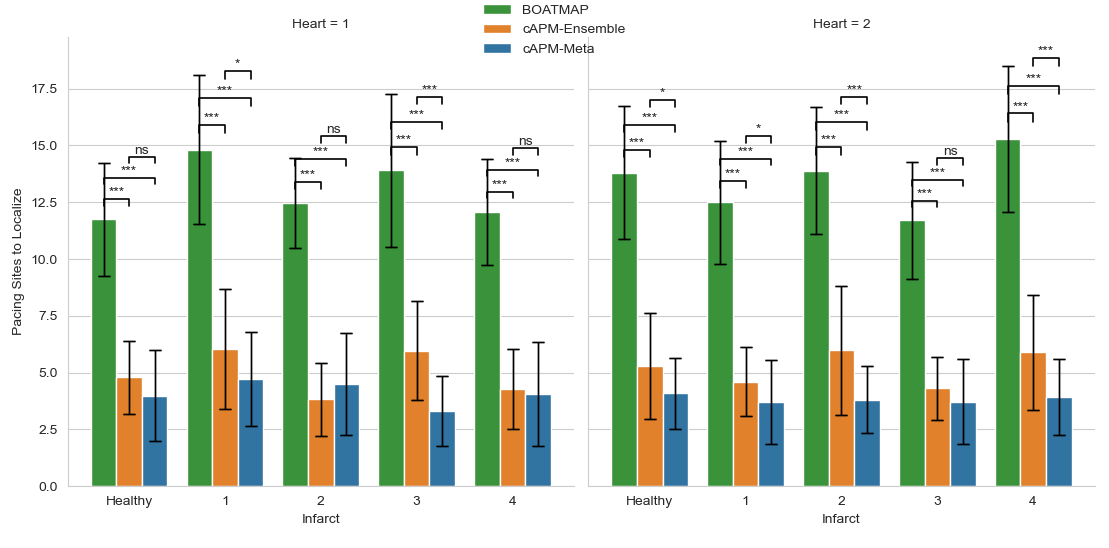}
  \caption{Summary of number of pacing-sites needed for localizing the target, averaged across 10 random seeds for a 12-task sequence for each heart and physiological conditions. Results show that both \ours methods vastly outperformed BOATMAP ($13.2 \pm 3.21$ pacing sites, while \ours-Meta ($3.98 \pm 2.10$ pacing sites) performed slightly better than \ours-Ensemble ($4.63 \pm 1.68$ pacing sites). $***$ indicates a statistically significant test of difference with $p <0.001$.}
  \label{ssbar}
\end{figure}

This advantage of \ours methods was consistently observed across all task sequences tested on both ventricles and all physiological conditions, as summarized in the number of pace-mapping required to localize shown in Figure~\ref{ssbar}. Compared to BOATMAP, both \ours methods needed a significantly smaller number of iterations to localize the target pacing site , confirmed by statistical test of significance (with $*** = p <0.001$). Furthermore, the number of pace-mapping sites needed were much more stable in the two \ours methods across tasks, demonstrating the benefits of knowledge transfer. Between the two \ours methods, \ours-Meta outperformed \ours-Ensemble on four out of the 10 settings (three with $* = p <0.05$, one with $*** = p <0.001$). When aggregating across all of the available data, \ours-Meta ($3.98 \pm 2.10$ pacing sites) performed slightly better than \ours-Ensemble ($4.63 \pm 1.68$ pacing sites), of which both vastly outperformed BOATMAP ($13.2 \pm 3.21$ pacing sites).

\begin{figure}[t!]
  \centering
  \includegraphics[width=\linewidth]{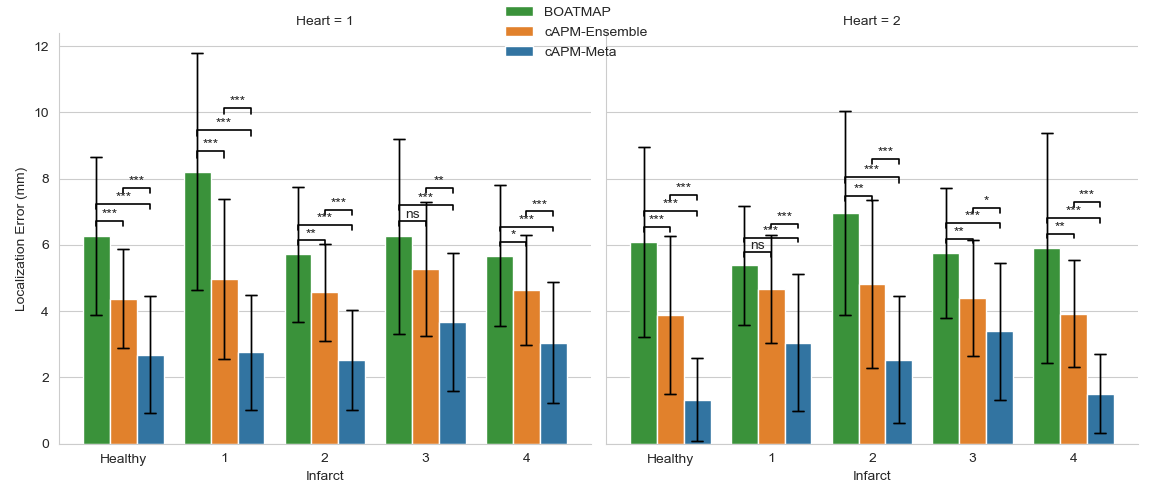}
  \caption{Summary of localization accuracy, following a similar presentation scheme as in Figure~\ref{ssbar}. Results show that \textit{cAPM-Meta}, \textit{cAPM-Ensemble}, and \textit{BOATMAP}  delivered a localization accuracy of $2.65 \pm 1.2$mm, $4 \pm 1.32$mm, and $4.9 \pm 2.5$mm, respectively.}
  \label{ss_accbars}
\end{figure}

\begin{figure}[t!]
  \centering
  \includegraphics[width=\linewidth]{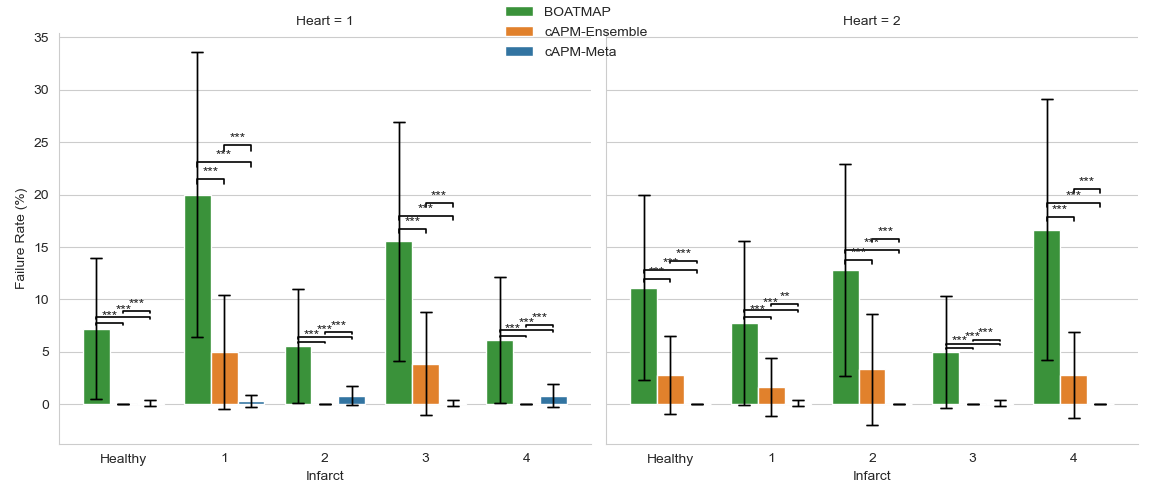}
  \caption{Summary of failure rates, following a similar presentation scheme as in Figure~\ref{ssbar}. Results show that \textit{cAPM-Meta}, \textit{cAPM-Ensemble}, and \textit{BOATMAP} had a failure rate of $0.7\%$, $1.8\%$, and $10.8\%$, respectively.}
  \label{ss_failurebars}
\end{figure}

\subsubsection{Localization Accuracy}
We also recorded localization error and failure rate as a measure of how precise and consistent each method is. 
The localization accuracy was calculated among the cases that a successful CC $\geq 0.97$ has been found with the target-ECG, thus an average of $\leq 5mm$ can be expected regardless of the methods used. The combination with failure rate, which measures percentage of cases that failed to find a target within 25 pace-mapping sites, thus provide a more complete assessment of the clinical utility of the considered methods in improving pace-mapping. Aggregating across all of the data within this setting, we find the localization error and failure rate of \textit{cAPM-Meta} $2.65 \pm 1.2$mm and $0.7\%$, of \textit{cAPM-Ensemble} $4 \pm 1.32$mm and $1.8\%$, and of \textit{BOATMAP} $4.9 \pm 2.5$mm and $10.8\%$. Both \textit{cAPM} methods outperform \textit{BOATMAP} significantly in both categories, but especially in the failure rate as anecdotally evidenced in Figure~\ref{ssovertime}. Interestingly, \textit{cAPM-Meta} has a localization error and failure rate almost half of \textit{cAPM-Ensemble} despite the similarity in number of pacing sites to localize.

Building on this, we can frame performance in terms of clinically actionable success by incorporating both the distribution of localization error and the observed failure rates. Specifically, if we assume the localization error for successful cases follows a Gaussian distribution parameterized by the reported mean and standard deviation, we estimate the probability that each method achieves an error below the clinically relevant threshold of $5$ mm. For \textit{cAPM-Meta}, this gives $\Phi\left(\frac{5 - 2.65}{1.2}\right) \approx 97.5\%$, which, after accounting for its $0.7\%$ failure rate, yields an overall cohort-level success rate of approximately $96.8\%$. For \textit{cAPM-Ensemble}, we obtain $\Phi\left(\frac{5 - 4.0}{1.32}\right) \approx 77.6\%$, and incorporating its $1.8\%$ failure rate gives an effective success rate of approximately $76.2\%$. In contrast, \textit{BOATMAP} achieves $\Phi\left(\frac{5 - 4.9}{2.5}\right) \approx 51.6\%$, which decreases to approximately $46.0\%$ when incorporating its substantially higher $10.8\%$ failure rate. This analysis highlights a pronounced separation between methods, demonstrating that \textit{cAPM-Meta} not only minimizes average localization error but also provides substantially greater reliability in achieving clinically acceptable accuracy.

\subsection{Task Sequences on the Same Ventricle across Different Physiological Conditions} \label{same_heart_different_physiology}

\begin{figure}[t!]
  \centering
  \includegraphics[width=\linewidth]{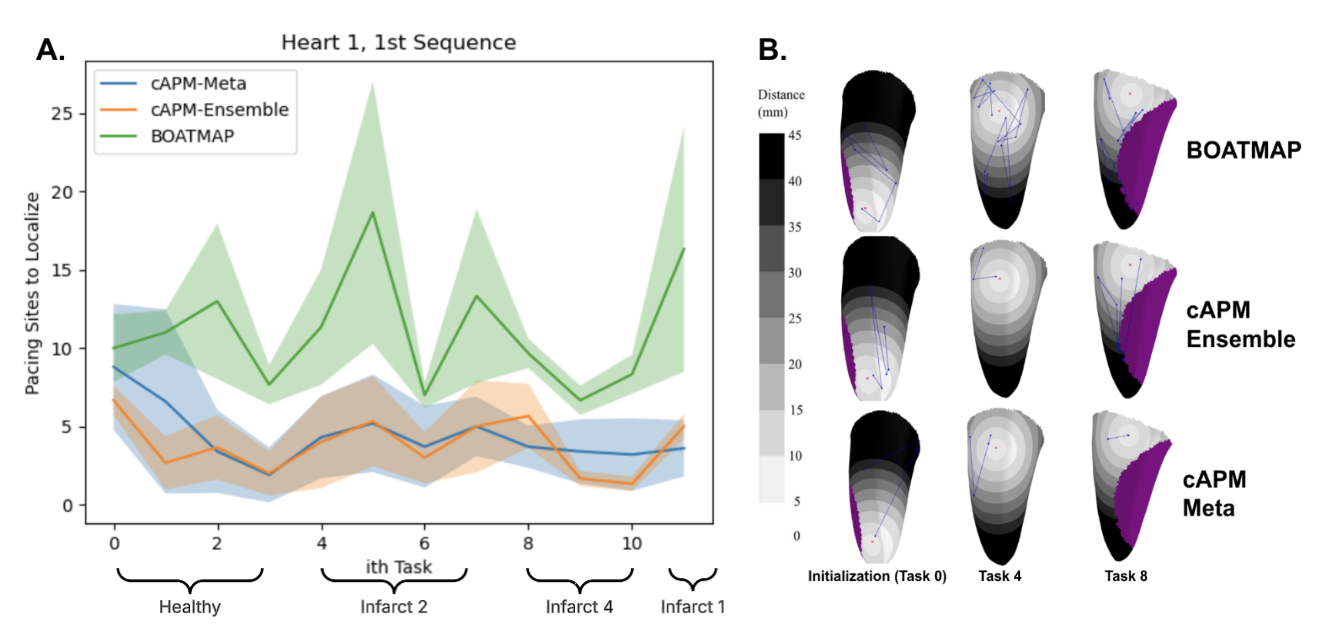}
  \caption{An example visualizing the number of pacing sites required by \ours-Meta, \ours-Ensemble, and BOATMAP to reach the convergence criteria on a sequence of 12 localization tasks on the same heart across different underlying physiological conditions. Note both the reduced number of pacing sites needed by both \ours methods and their stable performance across sequentially-presented tasks, including at task 4, 8, and 11 when the method was presented with localization tasks on scar settings that it has not yet seen in the past. B: Selected examples of localization tasks showing the number of pacing sites queried by BOATMAP \textit{versus} the two \ours methods.}
  \label{sd_overtime}
\end{figure}

\begin{figure}[t!]
  \centering
  \includegraphics[width=\linewidth]{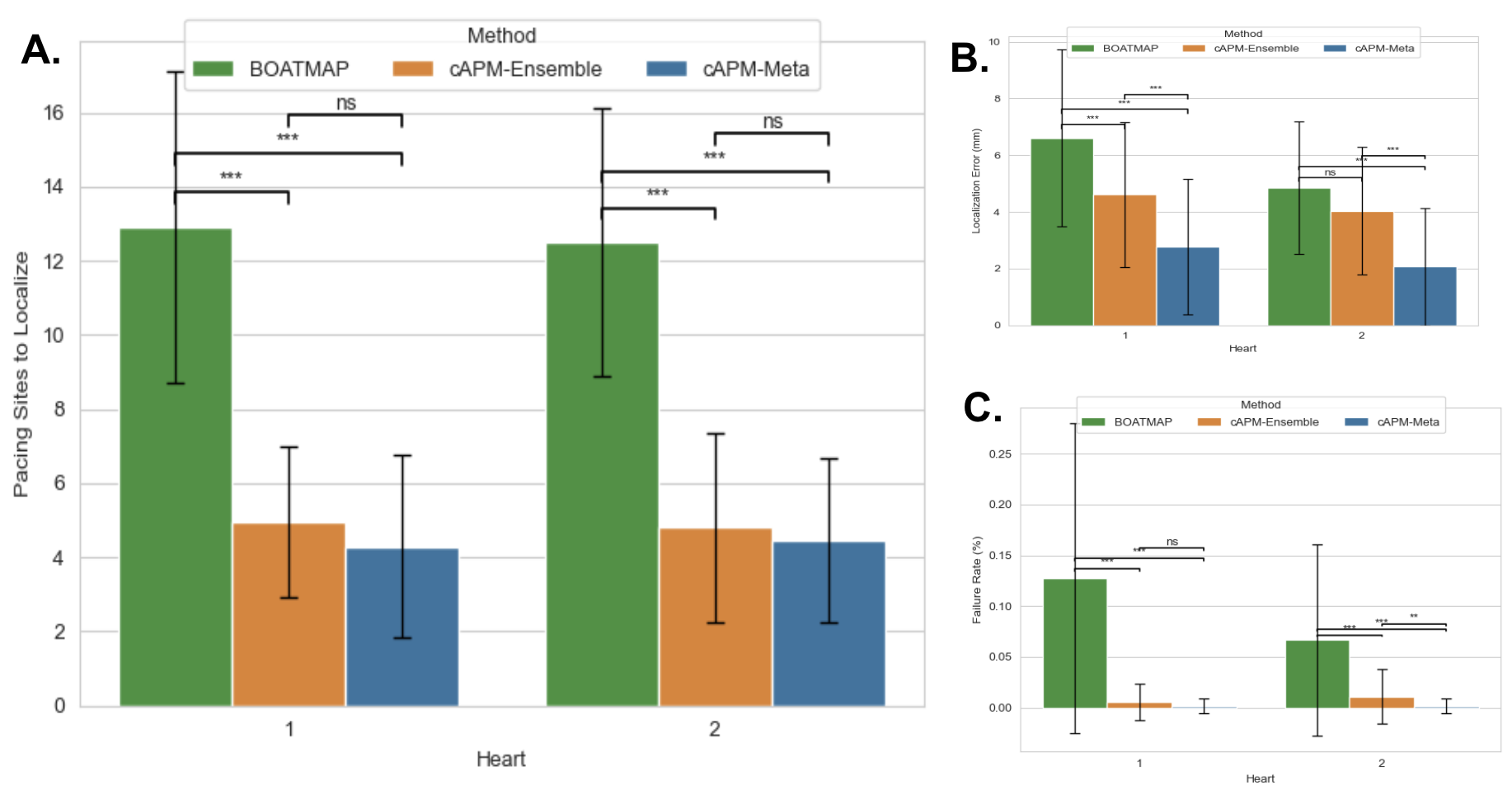}
  \caption{A. Summary of number of pacing-sites needed for localizing the target, averaged across 10 random seeds for a 12-task sequence (across healthy and various scar settings) on each heart. Results show that both \ours methods vastly outperformed BOATMAP ($14.8 \pm 3.21$ pacing sites, while \ours-Meta ($4.38 \pm 2.10$ pacing sites) performed slightly better than \ours-Ensemble ($4.88 \pm 1.68$ pacing sites). 
  B. Summary of localization accuracy and failure rates, following a similar presentation scheme as in A. Results show that \textit{cAPM-Meta}, \textit{cAPM-Ensemble}, and \textit{BOATMAP} had a localization accuracy and failure rate of $2.42 \pm 2.1$mm and $0.4\%$, $4.6 \pm 2.21$mm and $0.8\%$, and $6.1 \pm 3.72$mm and $13.2\%$, respectively.
  $***$ indicates a statistically significant test of difference with $p <0.001$.}
  \label{sd_combinedbars}
\end{figure}

In this set of experiments, we localized the pacing sites for different target ECGs on the same heart under \textit{varying} physiological conditions, such that the algorithm may see multiple different scar configurations within the same geometry. Clinically, this can be considered as attempting to localize across different patients with a similar ventricular geometry.

\subsubsection{Localization Efficiency}
Figure~\ref{sd_overtime} presents an example of the progression of all three models on a specific task sequence, the results across each metric we examined are very similar to the previous setting. For the first task, each model performs without significant difference to the other, but as the sequence progressed, BOATMAP continued to require 10-15 pacing sites to localize a target, whereas \ours required only 3-6 sites. One specific detail to note about this setting is the performance on tasks where the scar setting changed, in particular, on tasks 4, 8, and 11, where the algorithm has to localize a scar setting that has not yet been seen. Both \ours methods in this case are able to effectively transfer knowledge across scar settings, evidenced by the fact that there is no substantial decrease in performance on any of the relevant tasks, and they continue to show consistent improvement over the BOATMAP baseline.

This advantage of \ours methods was consistently observed across all task sequences tested on both ventricles, as summarized in the number of pace-mapping required to localize shown in Figure~\ref{sd_combinedbars} (left). Compared to BOATMAP, both \ours methods needed a significantly smaller number of iterations to localize the target pacing site, confirmed by statistical test of significance (with $*** = p <0.001$). Furthermore, the number of pace-mapping sites needed were much more stable in the two \ours methods across tasks, demonstrating the benefits of knowledge transfer. When aggregating across all of the available data, \ours-Meta ($4.38 \pm 2.10$ pacing sites) performed slightly better than \ours-Ensemble ($4.88 \pm 1.68$ pacing sites), of which both vastly outperformed BOATMAP ($14.8 \pm 3.21$ pacing sites). 

\subsubsection{Localization Accuracy}
We again recorded localization error and failure rate as a measure of how precise and consistent each method is. Aggregating across all of the data within this setting, we find the localization error and failure rate of \textit{cAPM-Meta} $2.42 \pm 2.1$mm and $0.4\%$, of \textit{cAPM-Ensemble} $4.6 \pm 2.21$mm and $0.8\%$, and of \textit{BOATMAP} $6.1 \pm 3.72$mm and $13.2\%$. Both \textit{cAPM} methods outperform \textit{BOATMAP} significantly in both categories, but especially in the failure rate as shown in Figure~\ref{sd_combinedbars} (right). However, as in the previous setting, we again observe \textit{cAPM-Meta} has a localization error and failure rate significantly less than that of \textit{cAPM-Ensemble}, showing there might be a tangible performance difference in terms of raw accuracy between the two \ours methods. 

Building on this same framework in the second setting, we again estimate the probability that each method localizes within the clinically relevant $5$ mm threshold. For \textit{cAPM-Meta}, $\Phi\left(\frac{5 - 2.42}{2.1}\right) \approx 89.0\%$, which, combined with its $0.4\%$ failure rate, yields an effective cohort-level success rate of approximately $88.7\%$. For \textit{cAPM-Ensemble}, we compute $\Phi\left(\frac{5 - 4.6}{2.21}\right) \approx 57.2\%$, which becomes approximately $56.7\%$ after incorporating its $0.8\%$ failure rate. \textit{BOATMAP}, by comparison, achieves $\Phi\left(\frac{5 - 6.1}{3.72}\right) \approx 38.4\%$, which declines further to approximately $33.3\%$ when adjusted for its $13.2\%$ failure rate.

\subsection{Task Sequences across Different Ventricles and Physiological Conditions}
\begin{figure}[t!]
  \centering
  \includegraphics[width=\linewidth]{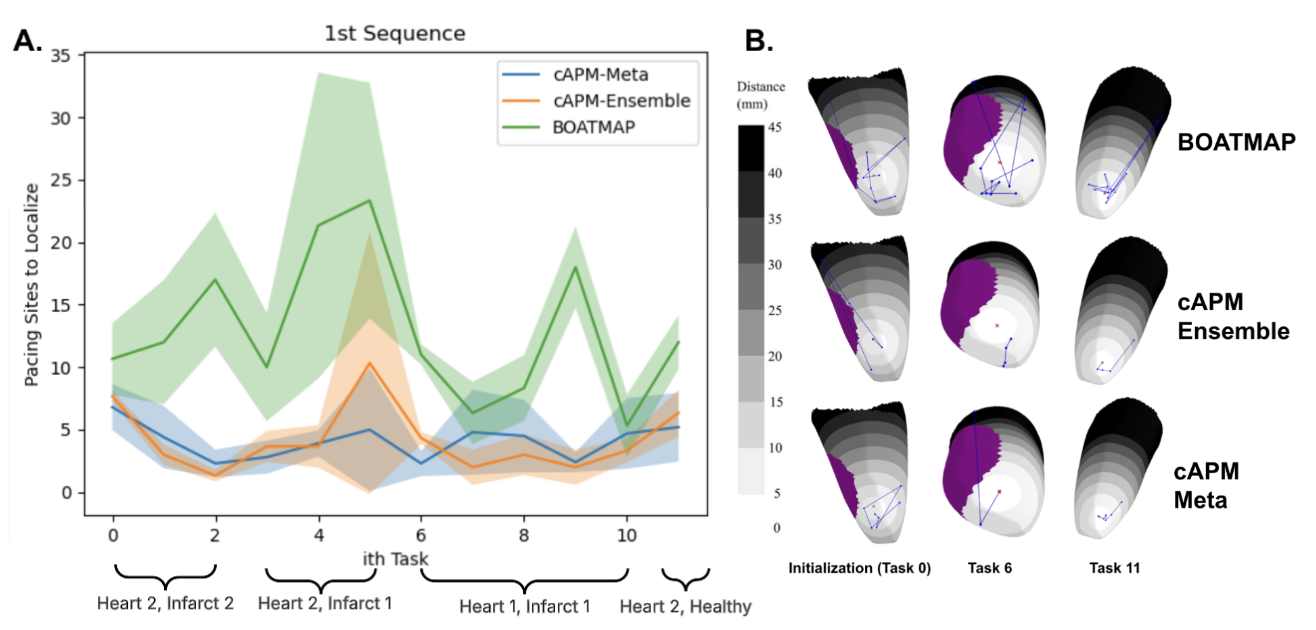}
  \caption{An example visualizing the number of pacing sites required by \ours-Meta, \ours-Ensemble, and BOATMAP to reach the convergence criteria on a sequence of 12 localization tasks across the two heart geometries and different underlying physiological conditions. Note both the reduced number of pacing sites needed by both \ours methods and their stable performance across sequentially-presented tasks, including at task 3, 6, and 11 when the underlying scar settings changed, and at task 6 and 11 when the underlying ventricular geometry changed.}
  \label{dd_overtime}
\end{figure}

\begin{figure}[t!]
  \centering
  \includegraphics[width=\linewidth]{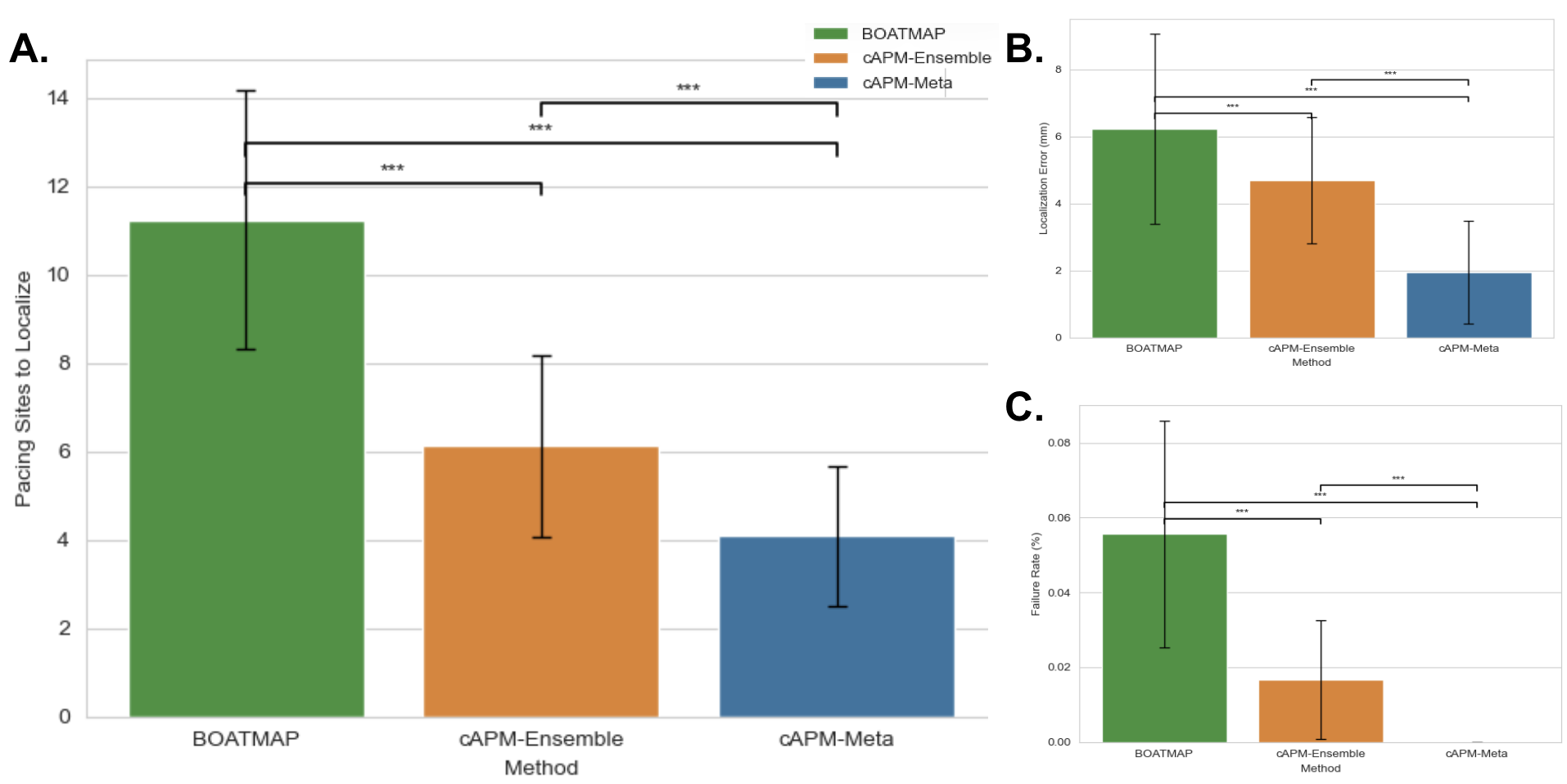}
  \caption{A. Summary of number of pacing-sites needed for localizing the target, averaged across 10 random seeds for a 12-task sequence across the two hearts and different physiological conditions. Results show that both \ours methods vastly outperformed BOATMAP ($13.5 \pm 3.21$ pacing sites, while \ours-Meta ($4.09 \pm 2.10$ pacing sites) performed slightly better than \ours-Ensemble ($5.01 \pm 1.68$ pacing sites). B. Summary of localization accuracy and failure rates, following a similar presentation scheme as in A. Results show that \textit{cAPM-Meta}, \textit{cAPM-Ensemble}, and \textit{BOATMAP} had a localization accuracy and failure rate of $1.94 \pm 1.7$mm and $0\%$, $4.5 \pm 1.9$mm and $1.6\%$, and $5.7 \pm 2.56$mm and $11.3\%$, respectively. $***$ indicates a statistically significant test of difference with $p <0.001$.}
  \label{dd_combinedbars}
\end{figure}

In this set of experiments, we localized the pacing sites for different target ECGs on varying heart and scar settings, meaning there is no guaranteed similarity between the tasks seen by our algorithm. Clinically, this equivalent to a real world scenario where many different patients with varying degrees of myocardial scaring undergo pace-mapping procedures.

\subsubsection{Localization Efficiency}
Figure~\ref{dd_overtime} presents an example of the progression of all three models on a specific task sequence. Again, we observed results very similar to both of our previous settings, indicating that our cAPM methods were efficiently transferring knowledge across heart geometries and scar configurations. Because the task sequences covered a broader variety of heart geometries and physiological conditions, BOATMAP's performance showed a larger variation compared to the previous settings, required on average 5-25 pacing sites to localize a target. In contrast, both \ours methods remained stable and continued to require only 3-6 sites once the sequence progresses. Note in particular at the tasks where the scar setting changed (3, 7, and 11) or even where the heart geometry changes (3 and 11): both \ours methods showed performance consistent with the other tasks, further confirming the effectiveness of the knowledge transfer techniques employed. Confirmed by statistical test of significance (with $*** = p <0.001$) as shown in Figure~\ref{sd_combinedbars}A, when aggregating across all of the available data, \ours-Meta ($4.09 \pm 2.10$ pacing sites) performed slightly better than \ours-Ensemble ($5.01 \pm 1.68$ pacing sites), of which both vastly outperformed BOATMAP ($13.5 \pm 3.21$ pacing sites). 

\subsubsection{Localization Accuracy}
Aggregating across all of the data within this setting, we found the localization error and failure rate of \textit{cAPM-Meta} $1.94 \pm 1.7$mm and $0\%$, of \textit{cAPM-Ensemble} $4.4 \pm 1.9$mm and $1.6\%$, and of \textit{BOATMAP} $5.7 \pm 2.56$mm and $11.3\%$. Both \textit{cAPM} methods outperform \textit{BOATMAP} significantly in both categories, but especially in the failure rate as shown in Figure~\ref{sd_combinedbars}C. We again observed a significantly smaller localization error in \textit{cAPM-Meta} compared to \textit{cAPM-Ensemble}. This trend was stronger in this setting than the previous two same heart settings, suggesting that knowledge transferring across geometries was more difficult and access to past data samples via the memory buffer in \textit{cAPM-Meta} was beneficial in this setting. Note that, however, \ours-Ensemble still met the need of the  clinical purpose of localizing a target within the error margin of $5$-mm (which was the size of a single ablation lesion); in contrast, while BOATMAP was able to deliver an average localization accuracy of $5.7$ mm, this was accompanied by a much larger failure rate.

Again we examine the performance of each method assuming a Gaussian distribution and determine the probability of localizing within clinical success ($< 5$mm). For \textit{cAPM-Meta}, this corresponds to $\Phi\left(\frac{5 - 1.94}{1.7}\right) \approx 96.4\%$, which, combined with its $0\%$ failure rate, yields an overall cohort-level success rate of approximately $96.4\%$. For \textit{cAPM-Ensemble}, we obtain $\Phi\left(\frac{5 - 4.4}{1.9}\right) \approx 62.4\%$, and accounting for its $1.6\%$ failure rate results in an effective success rate of approximately $61.2\%$. In contrast, \textit{BOATMAP} achieves $\Phi\left(\frac{5 - 5.7}{2.56}\right) \approx 39.2\%$, which further reduces to approximately $34.7\%$ when incorporating its substantially higher $11.3\%$ failure rate.

\section{Conclusions and Discussion}

In this work, we introduced \ours, with two technical variants depending on the underlying constraints on access to past data samples, for sequentially-presented tasks of localizing the origin of ventricular activation across geometries and physiological conditions. By integrating AL with either meta-experience replay or dynamically grown ensembles, both methods enable a general-purpose surrogate model to leverage knowledge from prior tasks rather than relearning from scratch for localizing a new target ECG. 

The experiments across increasing levels of task independence provided interesting insights into the robustness of knowledge transfer. In the simplest setting (same heart geometry and scar physiology), both \ours methods quickly exploited shared structure, leading to significant improvements. More interestingly, in the second and third settings where tissue properties and even heart geometries varied, performance remained stable. This indicates that the learned and transferred knowledge does not just capture local structures, but global ones as well. It is reasonable to expect our \ours strategies to work well within the same heart/scar configuration, but to see similar improvements in the cross heart/scar configurations is very promising. 

Another notable outcome is the reduction in failure rates, particularly for cAPM-Meta. BOATMAP exhibited failure rates exceeding 10\% in several settings, whereas cAPM-Meta reduces this to near zero. This is significant because failure cases in the pace-mapping setting is a particularly negative outcome in clinical practice. The inclusion of continual learning in \ours not only accelerated convergence but also improved the robustness of the search process. These results demonstrated that \ours can fundamentally enhance AL-driven pace-mapping in medical decision-making pipelines, enabling faster and more reliable localization with fewer interventions (\textit{i.e.}, localization within 5-mm error within 3-6 pace-mapping sites). These results, while obtained on the \textit{in-silico} testbed, provided strong empirical evidence to prepare the validation of \ours in \textit{in-vivo} settings both in preclinical and clinical studies, where pace-mapping decisions can be made on-the-fly as guided by \ours.

The broader implication of this work is that AL and continual learning are deeply complementary. AL focuses on selecting informative samples within a task, while continual learning ensures that information is retained and reused across tasks. The combination of the two yields a system that not only learns efficiently but also improves over time, approaching a form of lifelong learning tailored to clinical workflows.

\subsection{Knowledge Transfer by Access to Memory Buffer of Past Data Samples \textit{vs}. Past Surrogates}

Between the two methods considered (\ours-Meta and \ours-Ensemble), the distinction lies in how knowledge is transferred across tasks. cAPM-Meta performs data-level transfer via a memory buffer and gradient alignment, effectively learning a parameter initialization that is better suited for rapid learning. This mechanism appears particularly advantageous in terms of localization accuracy and failure rate. Across all settings, cAPM-Meta consistently achieved lower localization error and near-zero failure rates. This suggests that direct replay of representative past samples provides strong regularization and prevents drift in the learned surrogate, especially when tasks vary significantly. 

In contrast, cAPM-Ensemble performs model-level transfer, maintaining a collection of task-specific learners and combining them via a learned weighting function. This approach is more flexible in scenarios where data cannot be retained, but it did exhibit a modest degradation in accuracy relative to cAPM-Meta in our experiments. One plausible explanation is that the ensemble method relies on the weighter to infer task similarity implicitly, whereas cAPM-Meta directly exposes the model to past data distributions. Thus cAPM-Ensemble provides an informative, but noisy, prior representation of the ECG space, whereas cAPM-Meta is significantly more granular and detailed in its prior knowledge. Thus, cAPM-Ensemble can still converge quickly to a certain region within the heart, but may fail to exactly pinpoint the source of VT.

Further comparing our \ours methods there is a clear trade-off between memory and performance. cAPM-Meta requires maintaining a buffer of past samples, which may be constrained in real-world deployments due to privacy or storage limitations. cAPM-Ensemble addresses this by eliminating the need for data retention, but at the cost of slightly reduced accuracy. This trade-off is well-motivated, and the two methods together provide a spectrum of solutions depending on deployment constraints. Notably, the ensemble approach introduces its own complexity through the management of model growth and pruning, as well as the training of the weighting network. Future work could explore hybrid strategies that combine lightweight data summaries with ensemble representations to balance these considerations more effectively.

\subsection{Additional Ablation Studies}
\label{subsec:ablation}

\begin{figure}[t!]
  \centering
  \includegraphics[width=\linewidth]{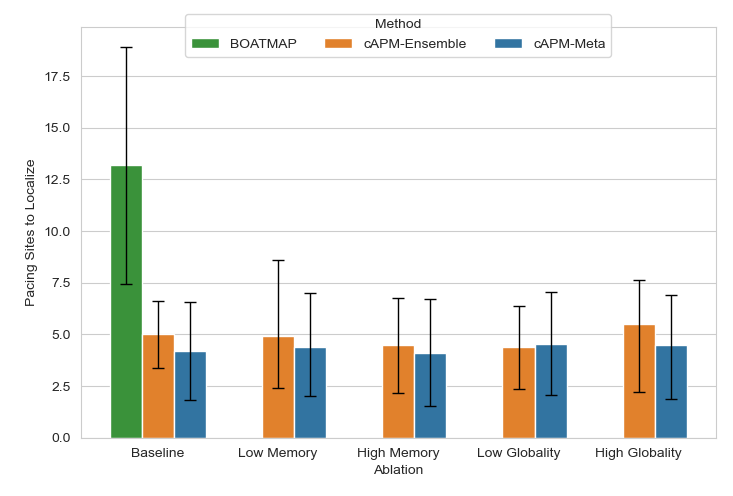}
  \caption{Summary of number of pacing-sites needed for localizing the target. Results show that across ablations each setting outperformed BOATMAP, where \ours-Meta showed little change in performance and \ours-Ensemble saw some improvements in increasing memory and reducing globality.}
  \label{ssbar}
\end{figure}

In longer sequences there may be more varied behaviors in our methods once the memory and continual learning strategies become stressed. Thus we tested our \ours methods on a sequences of $24$ tasks varying heart geometry and scar physiology. However, our results show performance almost identical to that of when testing over $12$ tasks. The BOATMAP baseline used $13.1 \pm 5.73$ pacing sites, cAPM-Ensemble used $5.04 \pm 1.55$ pacing sites, and cAPM-Meta $4.09 \pm 2.38$ pacing sites.

Additionally we conducted studies varying some of the key parameters for the \ours methods, the first being the size of allowed memory, and the second being the relative weighting between the global and local parts of the model. For the first parameter we analyze both a high and low setting compared to what was used for our central results. For cAPM-Meta this refers to the size of the memory replay bank: while $|M|$ was set to $10$ in our main experiments,  here we ablated $|M|=5$ and $|M|=20$. For cAPM-Ensemble, we altered the size of the ensemble $K$ from the value of $10$ used in the main experiments to values $5$ and $20$. When reducing the memory we see a slight performance degradation in cAPM-Meta ($4.38 \pm 2.38$) and little change in cAPM-Ensemble ($5.04 \pm 2.65$), but when increasing the memory cAPM-Ensemble performance improves significantly ($4.47 \pm 2.20$) while cAPM-Meta remains similar ($4.10 \pm 2.55$). The marginal improvement of performance of \ours-Meta when increasing the memory buffer size suggested that the model does not require a large set of data to extrapolate from, but rather a small diverse set of ECGs is sufficient for the model to create an accurate representation of the underlying truth. 

For the relative global/local weighting, we again analyzed both a high (more global) and low (more local) setting relative to those used in our main experiments. For cAPM-Meta, this corresponds to the meta learning rate $\beta$: while the main results used $\beta=0.003$, we additionally evaluated $\beta=0.0003$ and $\beta=0.03$. For cAPM-Ensemble, we varied the initial dropout rate $p$ from the baseline value of $0.5$ to $0.75$ and $0.25$. Altering the meta learning rate in cAPM-Meta degraded performance in both directions ($4.55 \pm 2.49$ and $4.5 \pm 2.53$). In cAPM-Ensemble, increasing the global weighting likewise reduced performance ($5.47 \pm 3.16$), whereas increasing the local weighting improved it ($4.37 \pm 2.01$), yielding a larger gain than that observed when increasing memory. These results suggest that the ensemble mechanism operates most effectively as a staged partnership, in the early active learning iterations, prior models provide useful global guidance, but after several unsuccessful rounds, the local model becomes more reliable on its own. At that point, reduced dependence on prior models appears beneficial, likely because those prior models were unable to provide sufficiently accurate guidance during the initial non-convergent iterations.

Overall, the change from hyperparameters resulted in some modest reductions or improvements in performance, where each configuration still showed vast improvements over BOATMAP in this stress-tested setting.

\subsection{Limitations and Future Works}

Experimental evaluations of AL-based technologies require prospective test settings where training samples are acquired on-the-fly based on AL, which is challenging and costly in biomedical settings for either human or animal-model testings. 
While the use of retrospective data is possible, previous works have shown that the sparse coverage of clinical pace-mapping data often results in unavailability of pace-mapping data at the sites suggested by the AL, 
hindering a rigorous evaluation \cite{missel2020hybrid}. 
The use of a high-fidelity \textit{in-silico} testbed addresses this challenge by providing high-resolution coverage of pace-mapping data that can be acquired by the AL at any locations needed. 
It however did not consider other practical factors, such as the accessibility of an AL-suggested pace-mapping site, the potential contact errors by the mapping catheters due 
to heart motion, 
or the decisions made by the clinicians. 
Our study however did lay down the foundation necessary to justify the cost and efforts for pursuing the next step of prospective preclinical and clinical validations of \ours.

Another limitation of the current study was that experiments were considered on only three different unique ventricle models, although within each we have considered five distinct setting of tissue properties. 
On each setting, we also only considered the localization target from paced-ECGs, instead of an induced reentrant VT of clinical interest -- the latter is in part due to the difficulty of inducing reentrant VT in an \textit{in-silico} setting.
Both limitations will be addressed as we move to preclinical and clinical validations considering animal models and human subjects presented with reentrant VTs.

Finally, \ours is intended to focus on assisting the specific clinical procedure of pace-mapping, 
the specificity of which for identifying the target of reentrant VT is suboptimal since 
it is attempting to localize the exit rather than the isthmus of the reentrant circuit \cite{Hongo19,Stevenson09}. 
While pace-mapping can be more effective for focal VTs and premature ventricular contractions (PVCs), it is often confounded by proximity to conduction system with fusion and different exit paths when near or within scar tissue. 
While these confounding factors were not specifically analyzed in this study, 
future works will 
delve into these more focused investigations into the clinical potential of \ours 
in addressing these confoundings, \textit{i.e.,} identifying pace-mapping sites that reside within or in the border zone of scar, or  near the conduction system without the chance for fusing with the conduction system. 

\section*{Acknowledgment} 
This work was supported by National Institutes of Health (NIH) award number
R01HL145590 (contact PI Wang).
This research was funded in whole or in part by the Austrian Science Fund (FWF) [grant 10.55776/ESP592] and by the National Science Foundation [grant NSF 2529303] awarded to K.G. 

\bibliographystyle{elsarticle-num} 
\bibliography{main}







\end{document}